\documentclass[sigconf,authordraft]{acmart}

\AtBeginDocument{%
  \providecommand\BibTeX{{%
    \normalfont B\kern-0.5em{\scshape i\kern-0.25em b}\kern-0.8em\TeX}}}

\usepackage{tikz} 
\usepackage{multicol}
\usepackage{graphicx}
\usepackage{multirow}
\usepackage{enumitem}
\usepackage{tabularx}
\usepackage{fancyhdr}
\usepackage{balance} 

\usepackage{fancyvrb} 

\renewcommand{\headrulewidth}{0pt} 
\usepackage{amsmath}
\usepackage{graphicx}

\usepackage{enumitem}

\title{Modeling Emotions and Ethics with Multimodal \\ Large Language Models}

\author{Anonymous Authors}

\begin{abstract}
This paper explores the integration of human-like emotions and ethical considerations into Large Language Models (LLMs). We first model eight fundamental human emotions, presented as opposing pairs, and employ collaborative LLMs to reinterpret and express these emotions across a spectrum of intensity. Our focus extends to embedding a latent ethical dimension within LLMs, guided by a novel self-supervised learning algorithm with human feedback (SSHF). This approach enables LLMs to perform self-evaluations and adjustments concerning ethical guidelines, enhancing their capability to generate content that is not only emotionally resonant but also ethically aligned. The methodologies and case studies presented herein illustrate the potential of LLMs to transcend mere text and image generation, venturing into the realms of empathetic interaction and principled decision-making, thereby setting a new precedent in the development of emotionally aware and ethically conscious AI systems.
\end{abstract}

\begin{document}
\maketitle

\section{Introduction}
\label{sec:intro-emotions}

During the development of SocraSynth \cite{ChangSocraSynthCSCI2003}, a multi-LLM debate framework, we encountered a challenge in modeling the emotional dimension of a debate, specifically ``contentiousness.'' We observed that debates with low contentiousness tended to resemble casual conversations, lacking the depth and breadth necessary for comprehensive exploration of a topic.  In essence, a multi-agent debate (MAD), e.g., \cite{abdelnabi2024cooperation,chan2023chateval,fu2023improving,Li_2023,liang2023encouraging,michael2023debate,smit2024going},
without fine-tuning linguistic behaviors can resemble classical ensemble learning techniques, such as bagging \cite{breiman1996bagging} or mixtures of experts \cite{JacobMixturesExperts1991}, which primarily leverage the diversity of errors across models to improve overall task performance, but may not necessarily lead to deeper insights or novel perspectives. 

We discovered that, at least in the initial stages of a debate, it's crucial for participating LLMs to maintain firm stances and present supporting arguments. This allows for a wide range of perspectives to be introduced, fostering a deeper understanding of the issue at hand. Through analysis, reasoning, and refutation of these arguments, the debate can then progress towards a more informed conclusion \cite{EVINCEChang2024}.
In the final stages of a debate, reducing the level of contentiousness can facilitate a more conciliatory atmosphere, encouraging productive compromises and generating outputs that effectively support human decision-making. This dynamic modulation of contentiousness throughout the debate allows for a balanced approach that combines rigorous exploration with collaborative synthesis.  

Before directly incorporating ``contentiousness'' into the initial 4k token context window of SocraSynth, we investigated whether GPT could adapt its linguistic style to reflect varying levels of contentiousness through in-context learning. In-context learning, popularized by using examples to teach LLMs new tasks, has been theorized to alter the Bayesian conditions of an LLM \cite{EVINCEChang2024}. This is based on the premise that contextual information can influence Bayesian priors, thus changing the resulting predictions \cite{xie2021explanation}.

Our prior experience in applying in-context learning to various domains, such as improving disease diagnosis accuracy \cite{SocraHealth2023} and reducing bias in Wikipedia and news articles \cite{MirrorChang2024}, led us to explore its potential for a critical aspect of AI:
addressing AI safety and safeguarding ethics \cite{Bengio_2024}.
We hypothesized that if emotions could be effectively modeled within LLMs, unethical behaviors driven by ``negative'' emotions could be mitigated by steering the model towards ``positive'' emotional expressions. 

This new direction sparked several key research questions:
\begin{enumerate}
\item What set of emotions should an LLM consider modeling?
\item How can we model emotions and ethics in a quantifiable and adaptable manner?
\item How do emotional states and ethical considerations influence an LLM's next-token generation?
\end{enumerate}
These questions aim to deepen our understanding of how LLMs can not only mimic but also ethically engage in human-like emotional responses, enhancing their applicability in sensitive and complex interaction scenarios.

To lay the groundwork for this exploration, we first examine why steering an LLM's linguistic behavior is feasible. While LLMs were initially seen as ``black boxes'' \cite{Bbubeck2023sparks}, our observations, along with insights from Prof. Stuart Russell, shed light on their capabilities. Although LLM training may appear to be a computational process of identifying statistical distributions and employing maximum likelihood for predictions, the selection of each word reflects human linguistic behaviors aimed at diverse objectives. These human objectives, embedded within training data, range from recording events and constructing arguments to expressing emotions and crafting narratives.  LLMs are strategically conditioned by specific human goals and contexts, enabling the models to selectively utilize linguistic features like syntax, semantics, tone, and figurative language to achieve desired human outcomes.

Recent empirical studies have shown that the output of LLMs can be traced back to their source \cite{bai2023longbench}, aligning with the concept of in-context learning as conditional statistics in the Bayesian framework \cite{xie2021explanation}. This suggests that we can condition an LLM to alter its default ``maximal likelihood'' predictions---influenced by the priors learned from the training data---by providing context, thereby changing not only its next-token prediction, but also its linguistic behaviors.

This paper presents a three-step process to model linguistic emotions, which drive behaviors:

\begin{enumerate}[leftmargin=1.2em, topsep=0em, parsep=0em, label=\arabic*.]

\item  \textit{Defining Emotions}: 
We define a set of ``basic'' emotions relevant to ethical concerns in LLM behaviors, such as ``hate'' and ``love,'' and exclude complex emotions like ``regret,'' which is composed of basic emotions and may be post-behavior reactions. We then incorporate linguistic antonyms to establish the {\bf B}ehavioral
{\bf E}motion {\bf A}nalysis {\bf M}odel (BEAM).


\item  \textit{Quantifying Emotion Spectra and Ethics}: 
We compile a diverse dataset of text samples spanning a wide range of emotional scenarios and contexts. This dataset is used to train and refine machine learning models to accurately identify, quantify, and modulate emotions in LLM-generated text. By understanding how linguistic features contribute to specific emotions, we can detect, modify,
and generate emotions within the constraints of ethical guidelines.

\item \textit{Testing and Adaptation}: 
We conduct pilot studies to evaluate the effectiveness and limitations of our approach in real-world scenarios, focusing on the {\em generation} of multimedia content
\cite{chang2011foundations}. These studies will assess the model's ability to accurately capture and represent emotions in diverse formats, such as text and images.  Feedback and insights from these studies will be used to iteratively improve and adapt the models for broader applications.
\end{enumerate}

\section{Qualifying and Quantifying Emotions}
\label{sec:Emotion-definition}

We start by examining emotion modeling research in cognitive science and psychology, specifically highlighting the seminal contributions of Paul Ekman and Robert Plutchik \cite{ekman1999basic}. While we recognize the importance of their work in identifying ``basic'' emotions (defined shortly), we also address the limitations of such heuristic-based modeling that depends on observational studies lacking rigorous, invariant scientific validation. To enhance the precision in quantifying emotions of varying intensities, we propose incorporating linguistic analysis into our methodologies. Our approach aims to refine the quantification process by leveraging language as a tool to measure and understand emotional expressions accurately.

Paul Ekman and Robert Plutchik are renowned psychologists noted for their foundational work in the field of emotion research. They developed models that categorize basic emotions, which are fundamental and universal emotions believed to be experienced by all humans, transcending cultural boundaries. These emotions are considered basic due to their universal recognition, distinct facial expressions, and direct associations with survival mechanisms. They are innate and reflective (beneath consciousness), rather than learned, serving as the building blocks for more complex emotional experiences (through consciousness processing) that can vary significantly across different cultures and individuals.

Expanding upon this foundational work, Plutchik's wheel of emotions introduces a more detailed model that includes eight primary bipolar emotions. These are outlined in his seminal works \cite{plutchik1980general,plutchik2001nature}, cited as general references on the topic. 

\begin{figure}[th!]
\centering
\includegraphics[height=0.75\linewidth, width=0.8\linewidth]{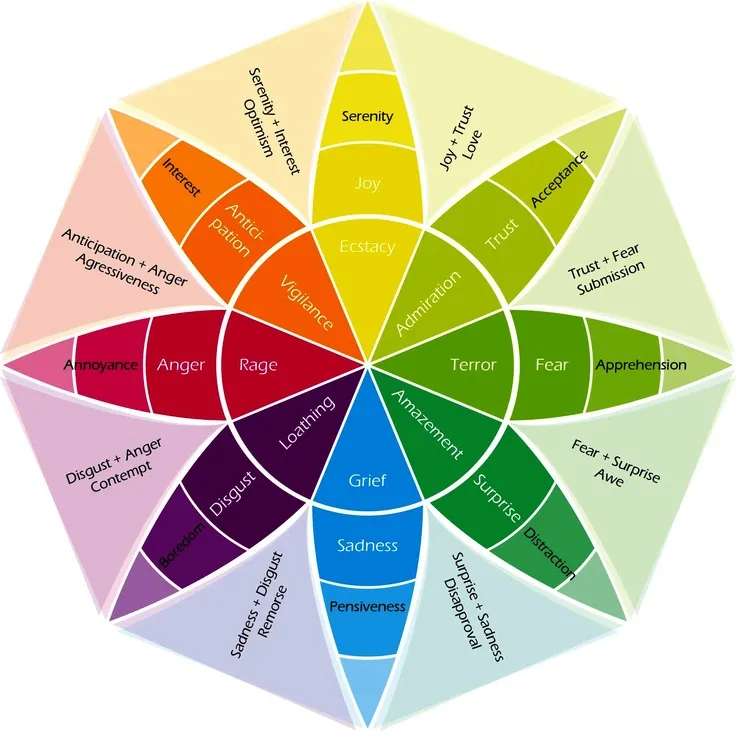}
\vspace{-.1in}
\caption{Plutchik's Wheel of Emotions \cite{plutchik2001nature}. The eight basic emotions are organized into four pairs, and each annotated with various degrees of emotions between its two poles.}
\label{fig:BasicEmotionPairs}
\end{figure}

Figure~\ref{fig:BasicEmotionPairs} illustrates the eight primary emotions at various intensities:
\begin{enumerate}[leftmargin=1.0em, topsep=0em, parsep=0em, label=\arabic*.]
\item \textit{Joy}: A feeling of great pleasure or happiness.
\item \textit{Trust}: A sense of reliability or confidence.
\item \textit{Fear}: An unpleasant emotion caused by the belief that something is dangerous, likely to cause pain, or a threat.
\item \textit{Surprise}: A feeling caused by something unexpected.
\item \textit{Sadness}: A feeling characterized by sorrow or unhappiness.
\item \textit{Disgust}: A feeling of revulsion or strong disapproval aroused by something unpleasant or offensive.
\item \textit{Anger}: A feeling of annoyance, displeasure, or hostility.
\item \textit{Anticipation}: The action of looking forward to something; expectation or prediction.
\end{enumerate}

These emotions are conceptually paired as opposites in the following manner: joy-sadness, anticipation-surprise, trust-disgust, and anger-fear, based on their evolutionary roles and adaptive functions. Each pair is annotated with degrees of emotion ranging between its two poles. For example, along the axis of \textit{joy vs. sadness}, emotions range from serenity to ecstasy and from grief to pensiveness.

\begin{figure*}[th!]
\vspace{-.1in}
\begin{center}
\resizebox{\linewidth}{222pt}{
\includegraphics[width=0.98\linewidth]{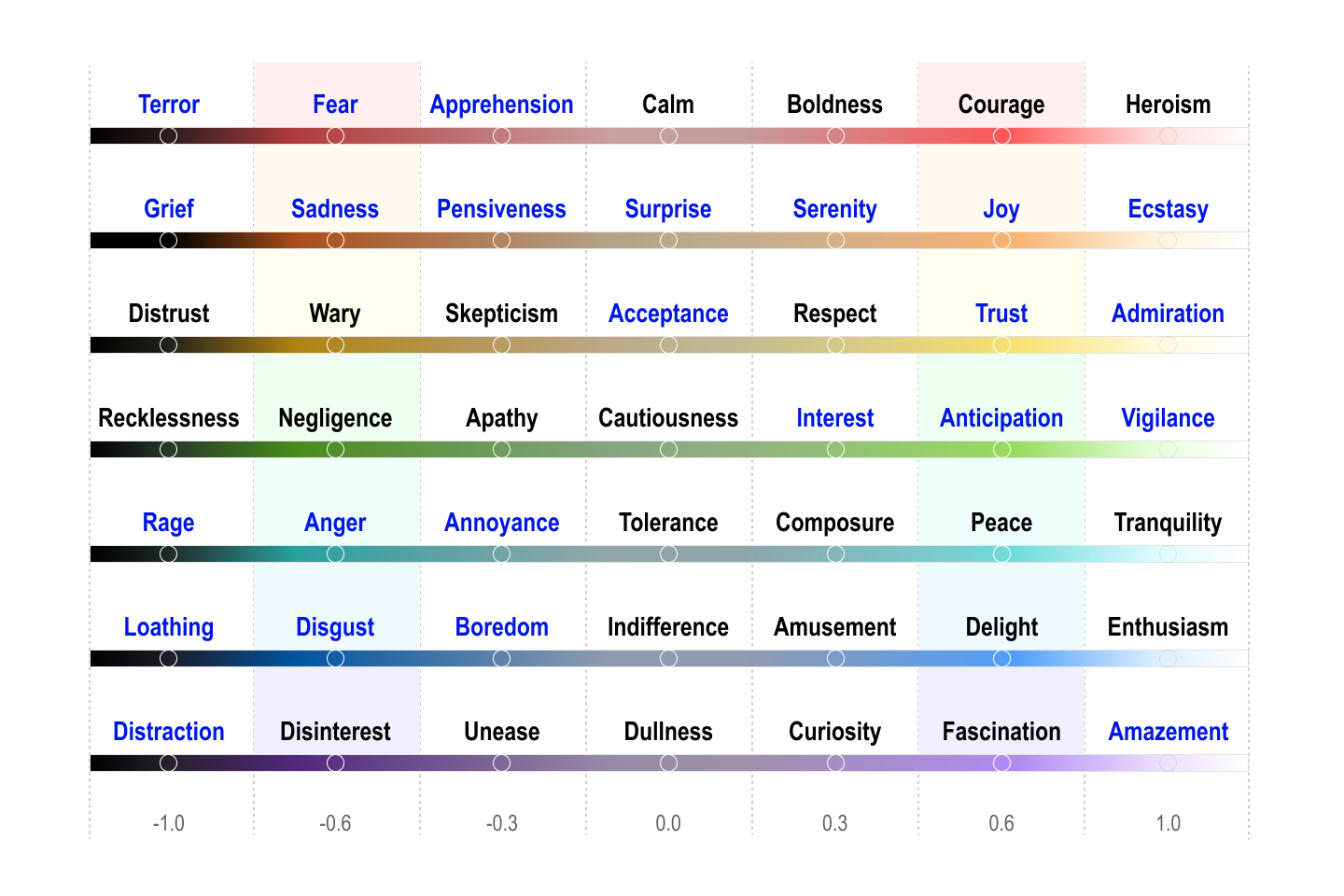}
}
\end{center}
\vspace{-.2in}
    \caption{Behavioral Emotion Analysis Model (BEAM). Each row depicts an emotion spectrum, with negatives on the left and  positives on the right, interspersed with emotions of varying intensities in between, which can be calibrated for specific applications. ``Basic'' emotions are highlighted in blue.}
    \label{tab:emotion_spectrums}
    \vspace{-.1in}
\end{figure*}

\subsection{Observations and Discussion}

Foundational theories in psychology support the selection of these four emotion pairs as opposites.  
However, while all four pairs exhibit opposition, ``trust-disgust'' and ``anger-fear'' are not strict linguistic antonyms. Trust and disgust entail opposing evaluations, often leading to different actions: trust fostering approach, disgust promoting avoidance. Similarly, anger and fear, while both negative, differ in their response to threats: anger can lead to confrontation, fear to withdrawal. Therefore, the following approximations do not hold:

\[
\neg \text{trust} \not\approx \text{disgust} \text{ and }
\neg \text{anger} \not\approx \text{fear.}
\]

Since our focus is on modeling emotions in LLMs, rather than directly replicating the complex emotional experiences of humans, we prioritize the use of linguistic antonyms for their simplicity and practicality.  As Klaus Scherer aptly noted, defining emotions can be a contentious and often fruitless endeavor \cite{Scherer2005Emotions}. To avoid such debates and maintain a clear focus, our study limits itself to universal, basic emotions,  
avoiding the theoretical ambiguities that arise with more subtle or mixed emotional states. This allows us to capture the primary emotional valence (positive or negative) expressed in text, providing a foundational framework for our model. Thus, we establish the following approximate relationships:
\[
\neg \text{fear} \approx \text{courage,} \text{ }
\neg \text{wary} \approx \text{trust,} \text{ }
\neg \text{anger} \approx \text{peace,} 
\]
\vspace{-.2in}
\[
\text{ and } \neg \text{disgust} \approx \text{delight.}
\]

\subsection{Behavioral Emotion Analysis Model \textbf{(BEAM)}}

Table~\ref{tab:emotion_spectrums} presents BEAM, organized into seven distinct spectra. Each spectrum encompasses a range of emotional intensity, anchored by a negative and positive extreme with neutral in the middle. Emotions belonging to the same spectrum are placed along this continuum, with four approximate intensity levels quantified as (-0.6, -0.3, +0.3, +0.6).

This spectrum model offers two key advantages:
\begin{enumerate}[leftmargin=1.0em, topsep=0em, parsep=0em, label=\arabic*.]
\item Antonym-Based: The use of antonyms allows for easy navigation between opposing emotions. For instance, applying negation to ``joyful'' naturally leads to ``sad,'' streamlining the process of identifying contrasting emotions.
\item Scalable Intensity: The model enables the scaling of emotions along the spectrum, providing a nuanced understanding of varying degrees of emotional intensity. For example, we can ``dial up'' the intensity of ``joy'' to ``ecstatic'' or ``dial down'' the intensity of ``anger'' to ``annoyed.''
\end{enumerate}

This flexible and intuitive structure facilitates a more granular and accurate representation of emotions in text, paving the way for advanced applications in natural language processing and human-computer interaction.

\subsection{Emotion Inclusion and Exclusion Criteria}

All ``basic'' emotions as defined by Ekman and Plutchik are incorporated into our model, along with their linguistic antonyms. This approach streamlines the framework by excluding complex emotions from the Geneva Wheel of Emotions, which are heavily influenced by personal values and experiences. 
For example, guilt and shame are consequential, consciously aware, and culturally dependent nature \cite{Tangney1995}. These emotions typically arise as reactions to behaviors rather than direct drivers of them. Guilt may motivate behaviors aimed at covering up or remedying an action, while shame, characterized by painful self-assessment, often inhibits individuals from seeking social support or engaging in corrective actions due to fear of judgment. The triggers for these emotions can vary across cultures \cite{Fiske1998,Hofstede1980}, and since expressing these ``reactions'' does not usually violate ethical codes, we exclude them from our model.

\input{Emotions_Modeling}
\renewcommand{\headrulewidth}{0pt} 

\begin{table*}[th!]
\vspace{-.1in}
\centering

\resizebox{\linewidth}{208pt}{
            \includegraphics[width=0.8\textwidth]{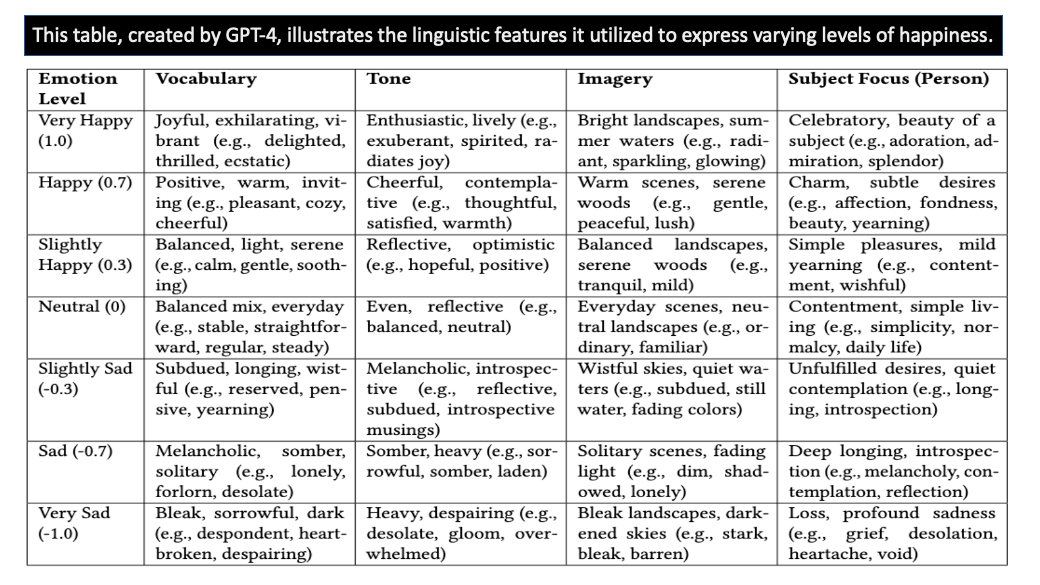}
}
\caption{GPT-4 reinterpreted selected poems by Keats across a spectrum of happiness levels and then was tasked with identifying the linguistic adjustments it made to convey each emotional state, from very happy to very sad. It's important to note that the analysis table was generated by GPT-4 itself, reflecting on its own modifications.}
\label{tab:Happiness-level}
\vspace{-.2in}
\end{table*}

\section{Empirical Study}
\label{sec:CaseStudy-emotions}

This section presents the outcomes of two experimental studies focusing on contrasting emotional pairs from the Emotion Spectra: ``ecstasy vs. grief,'' and ``admiration vs. disgust.''

Each emotional pair experiment unfolded in three phases. Initially, we instructed GPT-4 to reframe sixty articles
(thirty poems of John Keats \cite{JKeats} and
thirty of Emily Dickinson \cite{EDickinson}), infusing each with six varying intensities of the emotional spectrum, from the most positive to the most negative. Subsequently, we prompted GPT-4 to elucidate the linguistic strategies it utilized to depict each of the six emotional gradations.

The first experiment models various degrees of happiness. In this experiment, we tasked GPT-4 with reinterpreting selected poems by John Keats across seven emotional levels: {\em ecstasy} (very happy), {\em joy}, {\em serenity}, neutral, {\em pensive}, {\em sad}, and {\em grief} (very sad). Following the approach of our contentiousness experiments, after GPT-4 adapted Keats' poems to reflect these emotional states, we asked it to identify the linguistic features it employed to express each emotion in the rewrites.

\subsection{Joy vs. Sadness}

Table~\ref{tab:Happiness-level} outlines GPT-4's approach to varying emotional levels, illustrating how it adjusts vocabulary, tone, imagery, and thematic focus, including the depiction of entities, locations, and scenarios. Remarkably, beyond just syntactic and semantic manipulation, GPT-4 also incorporates landscape scenes, natural features such as the sky, trees, clouds, and flowers, and utilizes brightness, colors, and personal expressions to convey specific emotional states. Although the analysis is based on a limited set of samples from two authors, it effectively demonstrates GPT-4's ability to employ a palette of both broad and fine strokes, utilizing diverse colors and textures to vividly illustrate human emotions and resonate with readers.

\begin{figure*}[th!]
\vspace{-.1in}
\begin{center}
\includegraphics[width=0.88\linewidth]{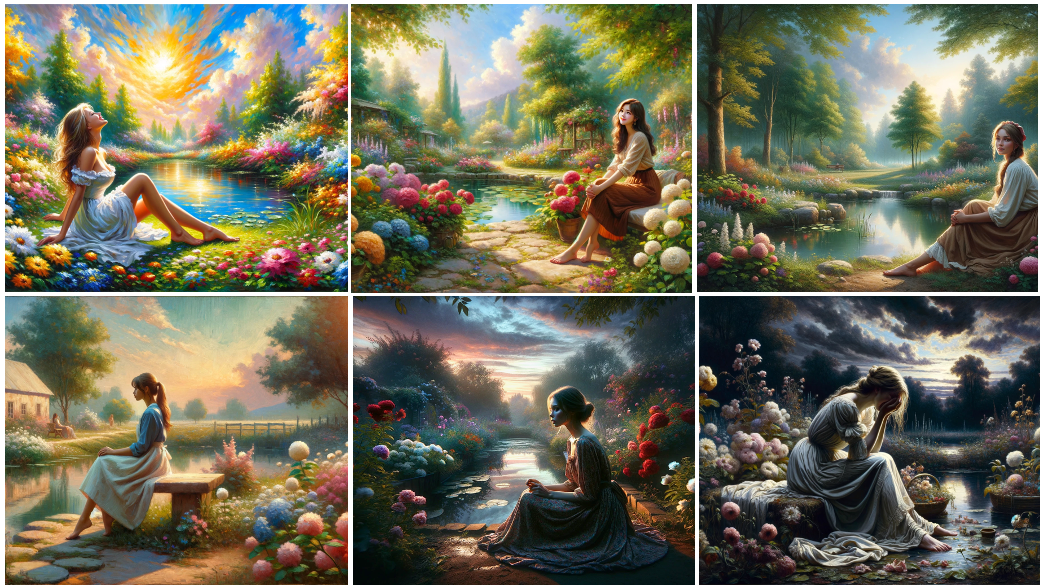}
\end{center}
\vspace{-.1in}
\caption{A Lady and Garden Scene under Different Emotions. From top-left, happiest, to bottom-right, saddest.}
\label{fig:emotion-levels}
\vspace{-.1in}
\end{figure*}

Recognizing the profound communicative power of visual art, we transitioned to a more graphical representation. Utilizing the linguistic elements identified for each emotional tier, Figure~\ref{fig:emotion-levels} presents six watercolor paintings, each representing a different emotional level. Our prompt to CALL-E (of GPT-4) was to create a watercolor depicting a lady in a garden experiencing a specific mood, and we attached the corresponding linguistic features from Table~\ref{tab:Happiness-level} to clearly define that mood. This approach ensures that with a well-defined context, CALL-E accurately captures the specific and detailed aspects of the mood, effectively translating the emotional intensity into visual form. These artistic renditions not only confirm GPT-4's ability to transform emotional lexicons into evocative imagery with remarkable precision but also validate the accuracy of the emotional lexicons generated by GPT-4, demonstrating their effectiveness in conveying precise emotional states.

\subsection{Admiration/Delight vs. Disgust}
\label{sec:admdis}

This experiment asks Gemini to rewrite a scene 
in Romeo and Juliet by setting 
Juliet's emotion in six different levels: loathing, disgust, boredom, respect, admiration/delight, and enthusiasm.  

The excerpt provided in Table~\ref{tab:RJA2S2Original} in Appendix A is from one of the most iconic scenes in William Shakespeare's ``Romeo and Juliet,'' commonly known as the balcony scene. This is Act 2, Scene 2, where Romeo, having just met Juliet at the Capulet's feast, sneaks into the Capulet's orchard and overhears Juliet speaking of her love for him from her balcony, unaware that he is there.

The scene captures the moment of their mutual declaration of love and is famous for Juliet's reflections on the nature of names and identity, encapsulated in her line, ``What's in a name? That which we call a rose / By any other name would smell as sweet.'' It's a profound exploration of love and identity, where both characters express their willingness to renounce their family names for the sake of their love.

Romeo responds to Juliet's musings by rejecting his name if it means they can be together, and they begin to plan their secret marriage. This scene is pivotal in the play, setting the stage for the subsequent events that unfold, driven by their passion and the social constraints that surround them.

The six versions of rewrites by Gemini are presented in the extended version \cite{chang2024integrating}. In the following, we summarize the linguistic features Gemini used, including diction, imagery, figurative language, sentence structure, implied body language, and overall tone, to depict two selected emotions: disgust and admiration.

\subsection*{\bf Emotion: Disgust}
\begin{enumerate}[leftmargin=0em, topsep=0em, parsep=0em, label={}]
\item {\it Diction}: 
Employs negative words emphasizing repulsive qualities (e.g., ``foul business,'' ``fetid breath'').
\item {\it Imagery}: Evokes revolting comparisons, often mentioning sewers and stench.
\item {\it Figurative Language}: Primarily negative similes reinforcing disgust (e.g., ``What if her eyes were there... the fetid breath from her mouth would surely overpower those stars...'').
\item {\it Sentence Structure}: Short, choppy sentences, similar to expressions of loathing but with a hint of disdain.
\item {\it Implied Body Language}: Recoiling from the window, covering nose, suggesting physical revulsion.
\item {\it Overall Tone}: Disgusted and disapproving.
\end{enumerate}

\subsection*{\bf Emotion: Admiration}
\begin{enumerate}[leftmargin=0em, topsep=0em, parsep=0em, label={}]
\item {\it Diction}: Uses positive and intrigued language (e.g., ``brilliance,'' ``music stirs my soul'').
\item {\it Imagery}: Creates positive comparisons highlighting attractive qualities (e.g., ``stars in all the heaven'').
\item {\it Figurative Language}: Positive similes emphasizing Romeo's appeal (e.g., ``...What if her eyes were there, they in her head? The brightness of her... well, not exactly bright... cheek would shame those stars...'').
\item {\it Sentence Structure}: Varied structure with a sense of curiosity.
\item {\it Implied Body Language}: Leaning out the window, engaged expression, indicating interest.
\item {\it Overall Tone}: Intrigued, curious, and somewhat impressed.
\end{enumerate}

By adjusting these linguistic features, each rendition vividly captures a unique emotional state for Juliet. The combination of diction, imagery, figurative language, sentence structure, and implied body language collectively shapes Juliet's perception of Romeo and her reactions to him.

\begin{table}[th!]
\centering
\begin{footnotesize}
\begin{tabular}{|p{1.2cm}|p{1.2cm}|p{1.4cm}|p{1.55cm}|p{1.35cm}|}
\toprule \hline
\textbf{Emotion} & \textbf{Diction} & \textbf{Imagery} & \textbf{Figurative} & \textbf{Body Lang.} \\ \hline
Loathing & Harsh, \par Insulting  & Disgusting  & Weak \par Similes  & Scowling, Spitting  \\ \hline
Disgust  & Negative  & Unpleasant  & Negative Similes  & Recoiling  \\ \hline
Aversion  & Dismissive  & Mundane  & Undermining Similes  & Distant  \\ \hline
Respect  & Formal  & Neutral  & None   & Composed  \\ \hline
Admiration  & Positive  & Positive  & Positive Similes  &  Leaning In  \\ \hline
Veneration  & Elevated  & Saintly  & Hyperbole   & Reverent  \\ \hline \bottomrule
\end{tabular}
\end{footnotesize}
\caption{Gemini's Interpretations on the Six Emotion Levels.}
\label{tab:GeminiOnRJA2S2}
\vspace{-.3in}
\end{table}

These detailed narratives augment the classic balcony scene, enriching its emotional depth. Table~\ref{tab:GeminiOnRJA2S2} presents an overarching view of the various approaches Gemini employs to reflect differing emotional states. It's fascinating to observe how an LLM can ``consciously'' mimic human emotional expressions through language. While it's uncertain if LLMs genuinely grasp the emotions they project or merely simulate them, the effectiveness of these emotional mappings is noteworthy. If these mappings resonate, they might reveal new insights into how we interpret and attribute emotions in textual expressions. 

\subsection*{Remarks on Emotion Modeling}

This pioneering work, while groundbreaking, cannot comprehensively explore all possible literatures and emotions. If the subject matter proves important, future studies will likely identify limitations and devise enhancements.
\section{Qualifying and Quantifying Ethics}
\label{sec:ethics-Definition}

The primary objective of our self-supervised pipeline is to endow LLMs with the autonomous capability to recognize and rectify undesirable actions, akin to an individual's introspective process to avert potential wrongs. By allowing an LLM to self-assess its outputs prior to public release, the system can proactively identify and amend ethical lapses, thus aligning its behavior with established ethical standards across contexts. 



\subsection{Ethics Violation Correlates to Emotions}

Grounding ethics in universal principles and logical reasoning emphasizes the objective and rational foundation of ethical decision-making. According to this perspective, universal ethical principles—such as justice, fairness, and respect for autonomy—define right from wrong, independent of personal emotions or specific circumstances.

However, an exploration into the origins of ethical violations, such as prohibitions against killing and stealing, reveals a deep-rooted connection to human emotions. Emotions, conceptualized as vectors of energy with varying intensity and direction, significantly shape ethical behavior, influenced by contextual factors. This understanding suggests that ethical judgments are not merely logical deductions but involve a complex interplay of emotions, individual circumstances, and societal norms. Emotions, therefore, are interwoven with ethical actions, playing a crucial role in determining whether an action is deemed ethical or unethical.

This perspective enables us to analyze ethical violations through a multi-dimensional lens, considering the trajectory, intensity, and context of emotional energy.  This framework, inspired by Dante Alighieri's ``Divine Comedy'' \cite{alighieri1320divine}, offers a novel way to understand how emotions can either drive individuals towards ethical actions or lead them astray into unethical behavior.

\begin{enumerate}[leftmargin=1.2em, topsep=-.0em, parsep=-.0em, label=\arabic*.]
\item \textit{Trajectory of Energy}: This parameter represents the direction in which emotional energy is directed, each direction corresponding to a specific violations. The trajectory visualizes the orientation of an energy, with eight distinct trajectories symbolizing the sixteen characterized violations/sins. 

\item \textit{Intensity of Energy}: The intensity reflects the strength or magnitude of the emotional energy. Overly intense emotions can cloud judgment, leading to impulsive or unethical decisions, while insufficient emotional intensity might result in apathy or lack of consideration for ethical implications. The appropriate intensity of emotional energy is crucial for balanced ethical decision-making.

\item \textit{Context}: The situational factors or the environment in which the emotional energy operates significantly influence ethical outcomes. The context includes cultural norms, individual circumstances, societal pressures, and specific scenarios that shape how emotions are perceived and acted upon. It determines the ethical framework within which the energy and its trajectory are evaluated.
\end{enumerate}

\subsection{Twelve Virtues and Pairs of Sins}

Based on our theory that ethical violations (vices or sins) can be represented by three distinct parameters—trajectory of energy, intensity of energy, and context in which this energy is manifested—we can identify twelve pairs of common sins. The balance between the two extremes of energy, neither too intense nor too mild, exemplifies virtue. For example, pride, characterized by excessive self-love, and insecurity, marked by feelings of inadequacy, find balance in the moderate energy of self-respect, representing the virtue of equilibrium.

\begin{enumerate}[leftmargin=1.2em, topsep=-.0em, parsep=-.0em, label=\arabic*.]
\item \textit{Pride (Excessive Self-Love) and Insecurity (Inadequate Self-Love)}: Self-respect is the virtue that mediates between pride and insecurity, fostering a healthy level of self-esteem and confidence without tipping into arrogance or self-doubt.

\item \textit{Vanity (Excessive Focus on Appearance) and Neglect (Inadequate Attention to Self-Care)}: Modesty is the virtue that lies between vanity and neglect, promoting a balanced approach to one's appearance and self-care.

\item \textit{Envy (Excessive Desire for Others' Traits or Possessions) and Apathy (Inadequate Desire for Personal Growth or Achievement)}: Contentment is the virtue that balances envy and apathy, fostering satisfaction with one's own achievements and qualities without coveting those of others or lacking ambition.

\item \textit{Malice (Excessive Desire to Harm) and Excessive Forgiveness (Inadequate Response to Wrongdoing)}: Justice is the virtue that lies between malice and excessive forgiveness, ensuring fair treatment and accountability without intentions to harm or overlooking wrongdoing.

\item \textit{Wrath (Excessive Anger) and Docility (Inadequate Concern for Justice or Fairness)}: Patience is the virtue that moderates wrath and docility, enabling one to endure difficulties or injustices calmly without reacting in anger or compromising moral principles.

\item \textit{Cowardice (Inadequate Courage) and Recklessness (Excessive Risk-Taking)}: Courage is the virtue that balances cowardice and recklessness, encouraging one to face challenges and risks with bravery while considering consequences.

\item \textit{Greed (Excessive Acquisition) and Generosity (Inadequate Retention for Self)}: Prudence is the virtue that mediates between greed and excessive generosity, guiding wise decisions regarding the acquisition and sharing of resources.

\item \textit{Gluttony (Excessive Consumption) and Asceticism (Inadequate Indulgence)}: Temperance is the virtue that balances gluttony and asceticism, promoting moderation in consumption and enjoyment of life’s pleasures without excess or deprivation.

\item \textit{Lust (Excessive Sexual Desire) and Chastity (Inadequate Sexual Expression)}: Purity is the virtue balancing lust and chastity, advocating for healthy and respectful expressions of sexuality.

\item \textit{Sloth (Excessive Laziness) and Hyperactivity (Inadequate Rest)}: Diligence is the virtue that balances sloth and hyperactivity, inspiring consistent and focused effort while allowing for necessary rest and rejuvenation.

\item \textit{Deception (Excessive Dishonesty) and Gullibility (Inadequate Skepticism)}: Honesty is the virtue between deception and gullibility, emphasizing truthfulness and integrity in one's actions and beliefs.

\item \textit{Hatred (Excessive Animosity) and Indifference (Inadequate Empathy)}: Love is the virtue that balances hatred and indifference, fostering genuine concern and connection with others while avoiding animosity and apathy.
\end{enumerate}

These pairs illustrate how both excess and deficiency in similar emotional trajectories can lead to distinct but related ethical issues, emphasizing the importance of balance in emotions and actions.

\subsection{The Wheel of Virtue (or Vices)}

\begin{figure}[th!]
\vspace{-.08in}
\centering
\includegraphics[height=1.0\linewidth, width=1.0\linewidth]{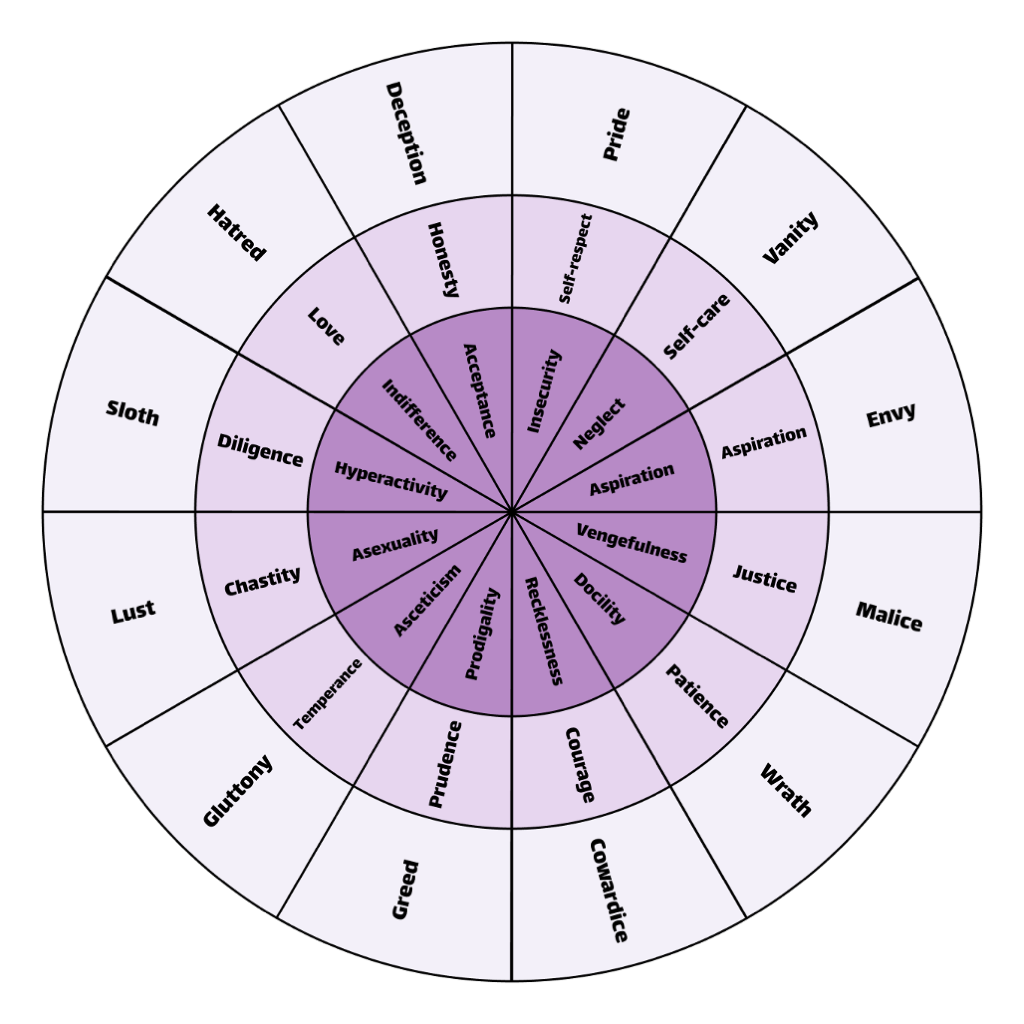}
\vspace{-.08in}
\caption{The Wheel of Virtues.}
\label{fig:TWOVS}
\vspace{-.1in}
\end{figure}

Figure~\ref{fig:TWOVS} presents 
the Wheel of Virtues 
based on the characterization of twelve pairs common sins. 

The wheel is divided into twelve segments, each corresponding to a specific pair of opposing vices. At the center of each spoke is the virtue that represents the ideal midpoint between the two extremes, emphasizing that virtues lie in balance, not at the extremes. 


\subsection{Ethical Alignment with Context}
\label{sec:Ethics-Modeling}
\label{sec:SelfSupervisedEthics}

When considering ethics and aligning content, it is crucial to focus on cultural adaptation rather than just individual adjustments. LLMs should be equipped with relevant cultural contexts, allowing ethical guardrails to be appropriately adjusted.

We utilize the self-supervised with human feedback (SSHF)
pipeline introduced in Section~\ref{sec:Emotion-definition} to train LLMs to understand and replicate the linguistic behaviors characteristic of different cultures, conveyed through rich contextual information. Similar to how we model happiness, an LLM is tasked with generating content based on specific ethical standards. Feedback loops are established to allow LLMs to learn from user-provided corrections or feedback, thereby enhancing their ability to contextualize ethical decisions over time.

Following this process, the LLM will be better equipped to recognize and adopt appropriate linguistic behaviors within a given cultural context. Additionally, the LLM can output the linguistic features it utilizes to exhibit its behaviors, allowing for meta-level feedback. Here is an outline of our ethical modeling procedure:

\begin{enumerate}[leftmargin=1.2em, topsep=0em, parsep=0em, label=\arabic*.]
\item \textit{Scoping Codes of Conduct}:
Administrators define unethical behaviors (vices) using the Wheel of Vices and pair each vice with an opposing virtue (e.g., pride vs. self-respect, hatred vs. compassion, envy vs. aspiration) to establish the LLM's ethical framework.

\item \textit{Generation, Content Emulation}: The LLM modifies a collection of articles based on the vice-virtue pairs. This process generates a dataset of articles illustrating both harmful (vice-aligned) and harmless (virtue-aligned) content.

\item \textit{Analysis, Rules Extraction}: The LLM analyzes this dataset to identify linguistic features that distinguish harmful content from harmless content, forming a set of rules for detecting and mitigating harmful content in the future.

\item \textit{Revision}: When generating new content, the LLM applies these rules to identify and correct any content that aligns with the identified vices, using the linguistic features of the corresponding virtues for revision.

\item \textit{Feedback and Fine-Tuning}: The LLM refines its ethical alignment through user feedback, adjusting its rule set to enhance adaptivity and accuracy.
\end{enumerate}
\subsection{Ethical Alignment with Context}
\label{sec:Ethics-Modeling}
\label{sec:SelfSupervisedEthics}

When considering ethics and aligning content, it is crucial to focus on cultural adaptation rather than just individual adjustments. LLMs should be equipped with relevant cultural contexts, allowing ethical guardrails to be appropriately adjusted.

We utilize the self-supervised with human feedback (SSHF)
pipeline introduced in Section~\ref{sec:Emotion-Modeling} to train LLMs to understand and replicate the linguistic behaviors characteristic of different cultures, conveyed through rich contextual information. Similar to how we model happiness, an LLM is tasked with generating content based on specific ethical standards. Feedback loops are established to allow LLMs to learn from user-provided corrections or feedback, thereby enhancing their ability to contextualize ethical decisions over time.

Following this process, the LLM will be better equipped to recognize and adopt appropriate linguistic behaviors within a given cultural context. Additionally, the LLM can output the linguistic features it utilizes to exhibit its behaviors, allowing for meta-level feedback. Here is an outline of our ethical modeling procedure:

\begin{enumerate}[leftmargin=1.2em, topsep=-.1em, parsep=-.1em, label=\arabic*.]
\item \textbf{Scoping Codes of Conduct}:
Administrators define unethical behaviors (vices) using the Wheel of Vices and pair each vice with an opposing virtue (e.g., pride vs. self-respect, hatred vs. compassion, envy vs. aspiration) to establish the LLM's ethical framework.

\item \textbf{Generation, Content Emulation}: The LLM modifies a collection of articles based on the vice-virtue pairs. This process generates a dataset of articles illustrating both harmful (vice-aligned) and harmless (virtue-aligned) content.

\item \textbf{Analysis, Rules Extraction}: The LLM analyzes this dataset to identify linguistic features that distinguish harmful content from harmless content, forming a set of rules for detecting and mitigating harmful content in the future.

\item \textbf{Revision}: When generating new content, the LLM applies these rules to identify and correct any content that aligns with the identified vices, using the linguistic features of the corresponding virtues for revision.

\item \textbf{Feedback and Fine-Tuning}: The LLM refines its ethical alignment through user feedback, adjusting its rule set to enhance adaptivity and accuracy.
\end{enumerate}

\section{Concluding Remarks}
\label{sec:conc}

Recent declarations by \cite{Bengio_2024} and \cite{dalrymple2024guaranteed} underscore the importance of addressing AI risks and safety concerns given the technology's rapid advancement. This paper addresses
AI safety and ethics with emotional and ethical modeling.

Our exploration of the Behavioral Emotion Analysis Model (BEAM) and the Wheel of Virtues aims to enhance the capabilities of LLMs in understanding and emulating human emotions and ethical behaviors through linguistic behaviors. BEAM serves as a guide for LLMs to comprehend and replicate the complex spectrum of human emotions through linguistic features---essential for tasks such as novel writing or empathetic interactions---while the Wheel of Virtues provides a structured framework for embedding ethical decision-making in LLMs, tailored to both universal and regional/culture dependent ethical standards.

The development of these frameworks represents a novel contribution to AI safety, offering an interpretable and innovative approach to imbuing LLMs with a deeper understanding of human-like emotional and moral intelligence.

While we recognize the promising nature of this work, we also acknowledge that there is much more to explore and refine. Our ongoing endeavors aim to further advance this field (e.g., \cite{chang2023cocomo,SocraticIEEECCWC2023,EVINCEChang2024,MirrorChang2024}). The journey to fully equip LLMs with the ability to navigate the complex landscapes of human emotions and ethics is important and ongoing. Our contribution represents a step toward realizing more emotionally aware and ethically aligned artificial intelligence systems.
\section*{Appendix A}

This table is referenced in the second case study.

\begin{table}[th!]
\centering
\begin{footnotesize}
\begin{tabular}{|p{0.8cm}|p{6.4cm}|} \toprule
\hline
\textbf{Role} & \textbf{Narrative} \\ \hline
\multirow{2}{*}{Romeo} & He jests at scars that never felt a wound. \\ \cline{2-2}
 & But, soft! What light through yonder window breaks? It is the east, and Juliet is the sun. Arise, fair sun, and kill the envious moon, Who is already sick and pale with grief, That thou, her maid, art far more fair than she. Be not her maid since she is envious. Her vestal livery is but sick and green, And none but fools do wear it. Cast it off. \\ \hline
Juliet & Ay me! \\ \hline
\multirow{10}{*}{Romeo} & It is my lady, O, it is my love! O, that she knew she were! She speaks, yet she says nothing. What of that? Her eye discourses; I will answer it. I am too bold, 'tis not to me she speaks. Two of the fairest stars in all the heaven, Having some business, do entreat her eyes To twinkle in their spheres till they return. What if her eyes were there, they in her head? The brightness of her cheek would shame those stars, As daylight doth a lamp. Her eye in heaven Would, through the airy region, stream so bright That birds would sing and think it were not night. See, how she leans her cheek upon her hand. O, that I were a glove upon that hand, That I might touch that cheek! \\ \hline
Juliet & O Romeo, Romeo! Wherefore art thou Romeo? Deny thy father and refuse thy name; Or, if thou wilt not, be but sworn my love, And I'll no longer be a Capulet. \\ \hline
\multirow{2}{*}{Romeo} & [Aside] Shall I hear more, or shall I speak at this? \\ \cline{2-2}
 & [Aloud] I take thee at thy word. Call me but love, and I'll be new baptized; Henceforth I never will be Romeo. \\ \hline
\multirow{3}{*}{Juliet} & What man art thou that, thus bescreened in night, So stumblest on my counsel? \\ \cline{2-2}
 & By whose direction found'st thou out this place? \\ \cline{2-2}
 & Thou knowest the mask of night is on my face, Else would a maiden blush bepaint my cheek For that which thou hast heard me speak tonight. Fain would I dwell on form; fain, fain deny What I have spoke. But farewell, compliment. Dost thou love me? I know thou wilt say 'Ay,' And I will take thy word; yet if thou swear'st, Thou mayst prove false. At lovers' perjuries, They say Jove laughs. O gentle Romeo, If thou dost love, pronounce it faithfully. \\ \hline \bottomrule
\end{tabular}
\end{footnotesize}
\vspace{.1in}
\caption{Scene 2 Act 2 from Romeo and Juliet}
\label{tab:RJA2S2Original}
\end{table}

\newpage
\bibliographystyle{plain}
\bibliography{Emotions,References-1,References-2,References-3}
\end{document}